# BioImage.IO Chatbot: A Community-Driven AI Assistant for Integrative Computational Bioimaging

**Wanlu Lei[1,2], Caterina Fuster-Barceló[3, 4], Gabriel Reder[5], Arrate Muñoz-Barrutia[3, 4], Wei Ouyang[5,*]**

[1]Department of Intelligent Systems, KTH Royal Institute of Technology, Stockholm, Sweden

[2]Ericsson Inc., Santa Clara, CA, USA

[3]Bioengineering Department, Universidad Carlos III de Madrid, Leganes, Spain

[4]Bioengineering Division, Instituto de Investigación Sanitaria Gregorio Marañón, Madrid, Spain

[5]Department of Applied Physics, Science for Life Laboratory, KTH Royal Institute of Technology, Stockholm, Sweden

*Correspondence: weio@kth.se

## Abstract

We present the BioImage.IO Chatbot, an AI assistant powered by Large Language Models and supported by a community-driven knowledge base and toolset. This chatbot is designed to cater to a wide range of user needs through a flexible extension mechanism that spans from information retrieval to AI-enhanced analysis and microscopy control. Embracing open-source principles, the chatbot is designed to evolve through community contributions. By simplifying navigation through the intricate bioimaging landscape, the BioImage.IO Chatbot empowers life sciences to progress by leveraging the collective expertise and innovation of its users.

# Main

Bioimaging is a rapidly evolving data-driven field that faces significant challenges due to the exponential growth in data volume and complexity. This expansion presents vast opportunities for discovery but also poses substantial hurdles for researchers who must navigate a complex ecosystem of sophisticated tools for analysis and acquisition, diverse technical documentation, and extensive online databases. Researchers utilize a range of acquisition software such as micro-manager[1], alongside bioimage analysis tools like ImageJ[2], Ilastik[3], and napari[4], complemented by specialized extensions and pipeline-focused Jupyter notebooks. Additionally, the field has wholeheartedly adopted the open science movement, leading to more coordinated efforts in developing community-based online indices and databases like the BioImage Informatics Index[5], the BioImage Archive[6], image.sc forum[7] and the BioImage Model Zoo[8]. While these community-driven initiatives provide broad access to a plethora of tools, studies, and AI models, they also contribute to a rich yet complex landscape.

Meanwhile, AI-driven conversational assistants, notably ChatGPT powered by GPT-based Large Language Models (LLMs)[9], have emerged as transformative technologies across multiple sectors. Recent advancements, such as the introduction of GPT-4[10] and multi-modal foundation models with vision capability[11], have further enhanced their capabilities. These methods and tools have streamlined complex processes across various domains such as language learning, event planning and technical support. Promisingly, recent perspectives have specifically highlighted the potential of employing such chatbots in bioimaging tasks such as intelligent microscopy[12] and image analysis via GPT code generation[13].

The BioImage.IO Chatbot emerges as an intuitive tool crafted to streamline computational bioimaging, and aims to support the entire spectrum of bioimaging tasks—from navigating technical documentation and databases, to generating code for data analysis, and using sophisticated microscope control and analysis interfaces. It works by harmonizing the strengths of LLMs, Retrieval Augmented Generation (RAG)[14], and an array of other advanced features such as code generation, function calling for invoking external tools and vision capability for visual inspection. The chatbot is built upon the ReAct framework[15] which combines analytical reasoning with precise action, facilitated by the extensibility offered through the ImJoy[16] plugin ecosystem and the Hypha service framework (which is a supporting server component in the Bioimage Model Zoo[8]). As a gateway to a vast, community-enhanced knowledge base and toolkit, the chatbot additionally boasts a collection of automated analysis tools and a dynamic platform for cultivating new functionalities (**Fig. 1**).

At its foundation, the chatbot leverages a repository curated and continually enhanced by the global community, hosted on GitHub. This dynamic, community-driven resource fuels the chatbot's RAG capabilities, facilitating the delivery of precise and context-aware answers from an extensive compilation of sources, as depicted in Figure 1a. These sources include detailed documentation from specific tools like ilastik[3], napari[4], and DeepImageJ[17], alongside databases such as ELIXIR bio.tools ([https://bio.tools/](https://bio.tools/)) and the ImageJ Wiki ([https://imagej.net](https://imagej.net)). By segmenting text sources into chunks and generating language embeddings for each segment, which are then stored in a vector database, we enable efficient similarity searches (**Methods**). This process matches the user's query with corresponding text embeddings, allowing us to use the nearest search results to enrich the chat assistant's context, leading to more precise responses. The ease of updating this database ensures that the chatbot remains aligned with the latest advancements in bioimaging. This approach not only keeps the chatbot current without the need for constant fine-tuning of the LLMs but also minimizes the occurrence of hallucinations in its responses.

More than just a conduit for information, the BioImage.IO Chatbot actively interacts with various online services and forums (**Fig. 1a**), including the BioImage Informatics Index (https://biii.eu) and the image.sc forum(https://image.sc) by constructing requests to remote servers automatically. Its extension-based architecture (**Methods**) facilitates the execution of AI models and analysis tools through code generation and execution in a in-browser Python code interpreter, broadening its utility across all user expertise levels.

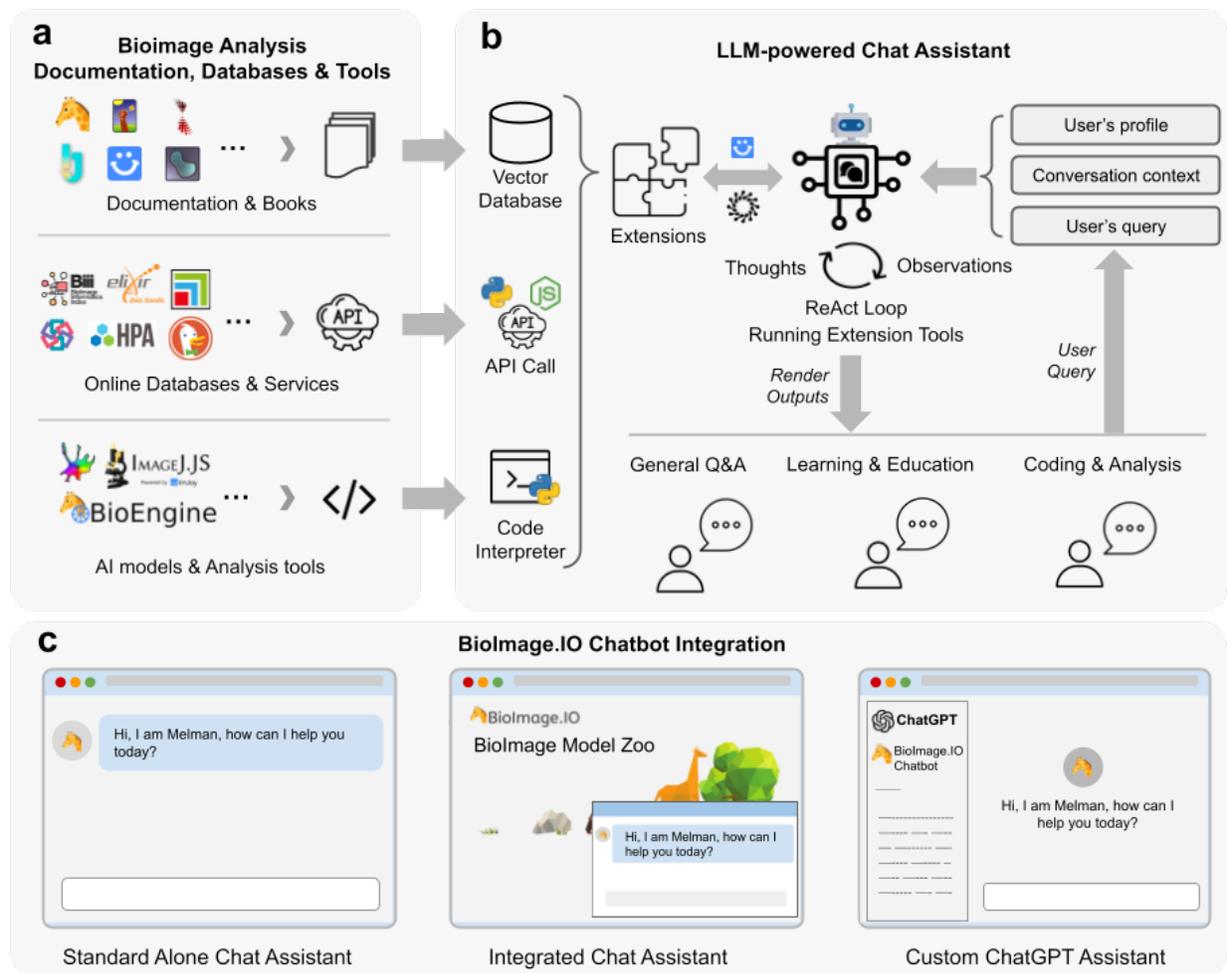

**Figure 1. Integration of BioImage.IO Chatbot for Enhanced Computational Bioimaging Support**
**(a)** The BioImage.IO Chatbot integrates a community-driven knowledge base for bioimage analysis, pooling from diverse resources including technical documentation, books, online databases, and AI-powered analysis tools. These resources inform the chatbot's responses, enabling nuanced support

> tailored to specific bioimage analysis inquiries. **(b)** At its core, the system amalgamates a vector database, API calls, and dynamic code generation within a code interpreter, linked through a unified extension mechanism. This facilitates a spectrum of user interactions, from basic inquiries to advanced educational assistance, code automation, and analysis execution. The system synthesizes user queries with context from the conversation and an optional user profile, leveraging a ReAct loop to generate logical sequences and action-specific responses. The assistant adapts to the query's complexity, providing instant answers or engaging extension tools for more elaborate assistance. **(c)** The BioImage.IO Chatbot is versatile in deployment, functioning as a standalone tool, an integrated feature within platforms like the BioImage Model Zoo, or in conjunction with OpenAI's GPT interfaces. It offers a standard "BioImage.IO Chatbot" GPT and the flexibility to create tailored versions, accommodating specific research needs.

As illustrated in **Fig. 1b**, the BioImage.IO Chatbot fuses the computational power of LLMs with a versatile extension mechanism to interact with a wide array of documentation, application programming interfaces (APIs), and resources. Tailored responses are crafted based on the users' queries, chat history, and profiles, enabling the chatbot to meet a broad spectrum of needs – from answering basic questions to providing educational support and executing complex analysis tasks. Central to its design is the ReAct loop, a dual-function system where LLMs generate reasoning traces and undertake specific actions in a synchronized manner (details in **Methods**). This synergy enhances the chatbot's ability to process complex inquiries and manage exceptions, allowing for refined search keywords based on the relevance of the current search results, search for documentation during coding and fixing code error based on the error stack trace (**Supplementary Video 2**). The chatbot adeptly navigates through intricate queries, accessing external resources to supplement its responses, thus offering rich, context-aware assistance to the bioimaging community targeting different user groups. The chatbot serves the bioimaging community with a range of functionalities, from engaging in straightforward Q&A and providing educational guidance to generating executable code for advanced image analysis.

The BioImage.IO Chatbot introduces three interaction modalities (**Fig. 1c**), each designed to enhance user access to its extensive toolkit. Users can engage with the chatbot directly for a wide array of informational and tool-based needs. Additionally, its integration as an ImJoy[16] plugin, particularly within the BioImage Model Zoo[8] (https://bioimage.io) and ImageJ.JS (https://ij.imjoy.io), showcases its versatility, allowing for seamless web ecosystem interactions. This integration leverages imjoy-rpc[16] for dynamic, in-context queries about models, datasets, and applications. In addition to the chatbot's direct and integrated applications, we offer a public API for our knowledge base and tools, compliant with the OpenAPI standard. This facilitates the development of OpenAI GPTs (customized ChatGPT), empowering users to create tailored chat assistants for inclusion in the GPT Store. By providing this API, we enhance our platform's openness and user focus, making complex computational bioimaging knowledge and tools more accessible to the wider research community.

The versatile nature of the BioImage.IO Chatbot is elucidated through a compendium of user scenarios powered by a series of extensions (**Methods**), demonstrating its utility in addressing a broad array of bioimaging challenges.

**Fig. 2** demonstrates the BioImage.IO Chatbot's versatility, showing its use in accessing documentation, querying online databases, running AI models for image analysis, developing extensions and controlling microscopes. The first three scenarios (**Fig. 2a-c, Supplementary Video 1**) illustrate users seeking insights within technical documentation or through online repositories, which underscores the chatbot's capacity to navigate its expansive knowledge base to deliver recommendations for bioimage analysis. This functionality is facilitated by its integration with vector databases and API calls, equipping it to provide informed recommendations tailored to bioimage analysis queries such as query cell images from the Human Protein Atlas[18] and image-related studies at the BioImage Archive[6].

Utilizing its code-generating, tool-calling, and vision capabilities, the BioImage.IO Chatbot exhibits advanced analytical and adaptive features through its ReAct loop mechanism. As illustrated in **Fig. 2d**, the chatbot utilizes its code generation capability and an in-browser Python code interpreter to process local images and run AI models via the BioEngine server[8]. In action, **Supplementary Video 2** showcases the chatbot's ability to emulate human programmers by iteratively generating code, executing it, analyzing results or errors, consulting documentation, and autonomously solving complex coding challenges. This process enables intricate image segmentation using tools such as Cellpose and the integration of newly developed libraries unfamiliar to the system. Additionally, GPT-4's vision capabilities not only enhance the chatbot's functionality by allowing for the evaluation of user images and analysis results, but also close the feedback loop. This feature provides real-time visual feedback, akin to that used by human bioimage analysts, enabling continuous performance improvement and a more interactive development cycle.

Furthermore, as depicted in **Fig. 2e**, the chatbot's capabilities can be expanded through its modular extension mechanism. Developers have the flexibility to engineer extensions within Jupyter Notebooks, or utilize ImJoy plugins and Hypha[8] services for browser-based or remote operations, respectively. With the ability to implement custom tool functions in Python or JavaScript, these extensions enhance the chatbot's functionality, enabling the automation of bespoke analysis workflows, seamless integration with existing web and desktop software, and hardware interaction for controlling microscopes[12,20,21] **(Fig. 2f)**.

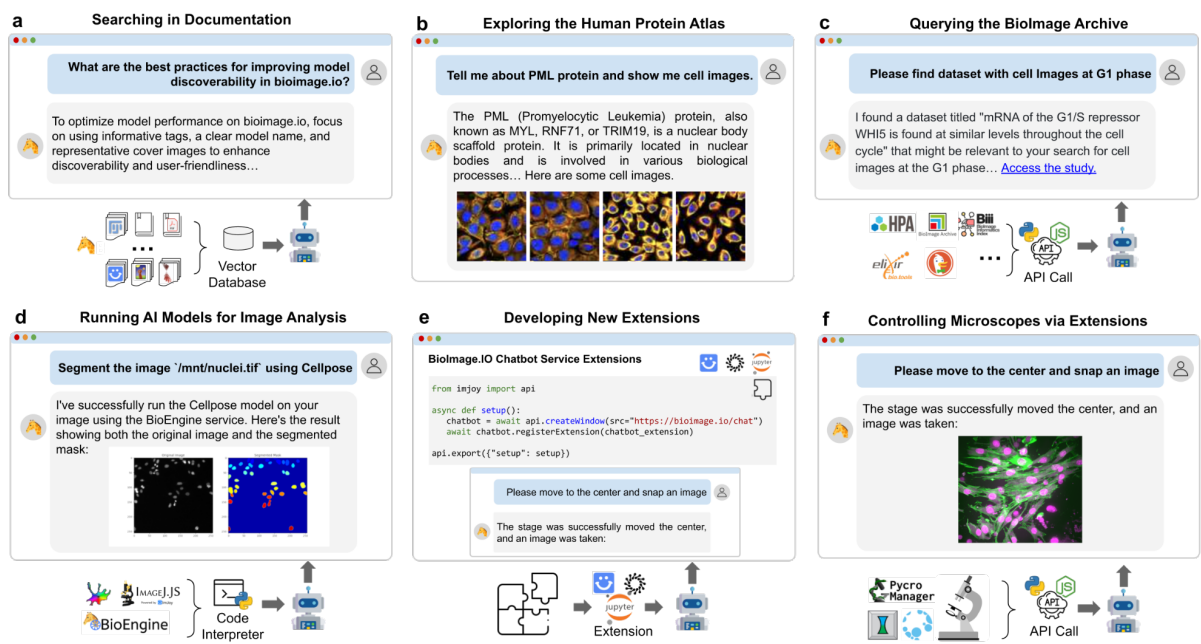

**Figure 2. Example Usage Scenarios of the BioImage.IO Chatbot: Querying Documents, Online Services, AI Model Execution, and Extension Development**

Features and user interaction capabilities of the BioImage.IO Chatbot: **(a)** Illustrates the chatbot's proficiency in navigating through and querying bioimage analysis documentation, efficiently extracting pertinent information from a vector database to address user inquiries about the BioImage Model Zoo community. **(b, c)** Showcases the chatbot's adeptness at exploring the Human Protein Atlas to understand the expression of proteins and accessing the BioImage Archive to retrieve relevant studies, utilizing API calls to integrate and deliver valuable findings from bioimage-related databases. **(d)** Captures the chatbot's capability to facilitate image analysis tasks by generating and executing Python code, it loads locally mounted user images, preprocesses it and sends it remotely to the BioEngine server to run the Cellpose model for nuclei segmentation. **(e)** Shows an example for developing extensions for the chatbot in a Jupyter notebook running in the browser. The extensions can be implemented in Python or Javascript, and registered as ImJoy plugins or hypha services. **(f)** Provides an example of extending the chatbot's functionality for controlling a microscope to move stages and acquire images.

Our commitment to the development of the BioImage.IO Chatbot is grounded in the principles of open-source development, emphasizing transparency, community engagement, and continuous improvement. We are acutely aware of the challenges presented by LLMs, including poor reproducibility, biases and hallucinations. In response, we adopt a collective approach to mitigate many of these issues. By employing RAG and vision feedback alongside a suite of curated tools, our strategy ensures that the chatbot sources information from reputable databases, technical documents and tools. This methodology substantially diminishes the risks typically associated with LLMs, thus providing an improved AI-assisted experience. Additionally, while we exclusively used proprietary LLMs from OpenAI, we acknowledge their limitations and maintain an adaptable system architecture for future transition to open-source LLMs. Our extension-based open design empowers the bioimaging community to directly drive the chatbot's enhancement and development. By inviting contributions, we leverage the collective intelligence of the community to foster the chatbot's continuous evolution, focusing on enhancing its accuracy and overcoming inherent LLM limitations. As we advance, the chatbot is poised to progress in alignment with community feedback, ensuring its position as a pivotal tool at the forefront of life sciences.

In sum, our efforts signify not just the construction of a digital assistant but the fostering of an inclusive community that shapes the future of computational bioimaging through shared knowledge and innovation.

## Data availability

Data used for supporting the BioImage.IO Chatbot are either publicly accessible from their original source or provided via the chatbot program https://github.com/bioimage-io/bioimageio-chatbot.

## Code availability

The source code for the BioImage.IO Chatbot is available at https://github.com/bioimage-io/bioimageio-chatbot.

## Acknowledgments

We thank Matus Kalas for his advice on accessing the bio.tools metadata and Curtis Rueden for his suggestions on accessing ImageJ wiki documentation and image.sc forum. We thank Matthew Hartley for providing guidance on accessing the BioImage Archive API and Hongquan Li for supporting the squid microscopy control demo. We appreciate the efforts of the AI4Life consortium members in creating and improving the BioImage Model Zoo and its documentation. We are also grateful to Florian Jug for testing our system and providing valuable feedback on its design. Last but not least, we thank Weize Xu for his kind support in implementing and testing chatbot extensions. This work was partially supported by the European Union's Horizon Europe research and innovation program under grant agreement number 101057970 (AI4Life project) awarded to A.M.B. and W.O., the SciLifeLab & Wallenberg Data Driven Life Science Program (grant: KAW 2020.0239) awarded to W.O. and by Ministerio de Ciencia, Innovación y Universidades, Agencia Estatal de Investigación, under grant PID2019-109820RB-I00, MCIN/AEI/10.13039/501100011033/, co-financed by European Regional Development Fund (ERDF), 'A way of making Europe' awarded to A.M.B. Views and opinions expressed are however those of the authors only and do not necessarily reflect those of the European Union. Neither the European Union nor the granting authority can be held responsible for them.

The authors utilized the language model ChatGPT developed by OpenAI to assist in structuring and drafting this paper.

## Author contributions

Conceptualization was led by WO, with supporting code and design contributions from WL, CFB, and AMB. The development and implementation of the BioImage.IO Chatbot were carried out by WL, GR and WO.

CFB organized the documentation, performed testing and user interaction design, while the manuscript was organized and written by CFB, WL, AMB, and WO. Funding and project administration were managed by AMB and WO.

## Competing interest

WO is a co-founder of Amun AI AB, a commercial company that builds, delivers, supports and integrates AI-powered data management systems for academic, biotech and pharmaceutical industries.

The remaining authors declare that the research was conducted in the absence of any commercial or financial relationships that could be construed as a potential conflict of interest.

# Methods

## BioImage.IO Chatbot Software Implementation

The chatbot software is a real-time application built upon the Hypha framework (https://github.com/amun-ai/hypha), incorporating a Python-based backend and a web frontend. The backend functionalities are split into several primary components: a knowledge base built using the FAISS vector database and a server mechanism for handling chatbot operations and its extensions. As described in the sections below, the knowledge base part involves downloading and parsing text from community-contributed documentation as listed in the manifest file, breaking this text into manageable chunks, creating embeddings for these chunks, and storing them in the FAISS database for quick retrieval.

In addition to its knowledge base functionalities, the backend also supports various assistants powered by GPT models through a Python library called "schema-agents." Each assistant is defined by a unique name, a set of instructions, and associated tools which are Python functions annotated and integrated through the chatbot's extension mechanism. This extension mechanism, detailed in another method section, allows for the incorporation of various functionalities such as the microscope control, which interfaces directly with microscope software via Hypha, and code execution within the user's browser using the Pyodide-based code interpreter extension.

The frontend of the chatbot handles user interactions, which include receiving input and displaying outputs from the backend, managing server communication, and facilitating the switch between different assistants or chatbot servers through URL queries. Extensions for the BioImage Model Zoo are implemented in the frontend using the ImJoy plugin framework, and additional online resources like the Human Protein Atlas and BioImage Archive are integrated as independent extensions. This comprehensive structure ensures that the chatbot can efficiently respond to documentation-related queries by leveraging its extension-based knowledge base and provide a versatile and interactive user experience.

## Community Knowledge Base

The community knowledge base is constructed from a curated list of data sources, maintained collaboratively on our GitHub repository. Contributors can submit pull requests to add new sources, which may be formatted as markdown or PDF files. In the process of building the database, files are first downloaded from the specified URLs. Subsequently, these documents are divided into manageable chunks, each containing approximately 1000 characters.

For every chunk of text, we employ OpenAI's text embedding technology to generate a corresponding vector representation. These vectors are then stored in a FAISS vector database, facilitating efficient similarity searches. Upon receiving a query, the chatbot utilizes the LLM to reformulate the query, which is then encoded using the same OpenAI embedding framework. By identifying the closest matches based on cosine distance among the precomputed embeddings, we can furnish the chatbot with contextually relevant information to inform its responses.

To ensure accessibility and ease of use, the vector database is automatically generated within our GitHub CI pipeline and subsequently uploaded to an S3 storage server. This allows users to readily access and incorporate the latest database updates without the need for local rebuilding, keeping the chatbot aligned with the latest developments in bioimaging research.

## ReAct Loop Implementation

The ReAct Loop is a central component of the BioImage.IO Chatbot, designed to enhance its response accuracy and adaptability. Upon receiving a user query, this mechanism initiates by aggregating various inputs, including the user's chat history, profile data, and information from a range of chatbot extensions. These extensions enable access to diverse data sources, from database queries to content retrieval from specialized forums such as image.sc.

The assistant processes this amalgamated data set through a selective mechanism, activating relevant tool functions tailored to the specific context of the query. It employs a combination of reasoning and task-specific actions, enabling a critical evaluation of the information retrieved by these tools. In instances where the initial response is inadequate or tool execution errors are detected, the system is designed to iteratively refine its approach. This iterative refinement process is critical for adjusting search parameters, rectifying errors, and incrementally enriching the response quality, akin to a human-like iterative problem-solving strategy.

The ReAct Loop persists until a contextually relevant response is formulated or a predetermined iteration limit is reached, ensuring a balanced approach between thoroughness and efficiency. Upon completion, the chatbot provides a detailed summary of the actions taken, preserving transparency throughout the interaction. Through this methodology, the ReAct Loop markedly improves the chatbot's operational efficiency, precision in problem-solving, and capacity to adapt to diverse user inquiries.

## Chatbot Extension Mechanism

The BioImage.IO Chatbot incorporates a modular extension mechanism designed to foster an open and collaborative development environment. This framework allows for the integration of native chatbot extensions developed in Python, utilizing Pydantic annotations or a "get_schema" function for defining input structures and validation rules for tool functions. Such design facilitates the accurate and seamless incorporation of tools into the chatbot's operation, enhancing its functionality and user interaction.

Furthermore, to embrace wider community participation, the chatbot supports client-side extensions via imjoy-rpc or server-side extensions via hypha, compatible with both JavaScript and Python. These extensions require a "get_schema" function for specifying JSON schema annotations, ensuring uniformity and precision in the chatbot's engagement with these tools. The support for dual-language extensions encourages diverse contributions, expanding the chatbot's functionality across a wide range of applications.

Client-side extension integration via imjoy-rpc offers simplified server-side management while decentralizing computational demands to the user's local environment. This architecture not only alleviates the server's computational burden but also enhances security by minimizing the risks associated with executing arbitrary code on server infrastructure.

Furthermore, developers can also serve chatbot extensions using hypha services in a distributed manner, this further increases the versatility of the extension systems, enabling hardware devices (e.g., microscope scope control) or computational resources (GPU) access to the extension. Since the communication is mediated with an external public server, it helps the user or developer to circumvent most of the network issues, allowing collaborative use of chatbot extensions. To further increase the utility of the extensions, we serve them over HTTP interface along with their schema in the OpenAPI standard. This allows users to create custom GPTs which interact with the extensions using OpenAI's GPT Creator. We provide a guideline on how to create a user's own GPT using our services in our GitHub repository. A prime example of this mechanism in action is the BioImage Model Zoo extension, which illustrates the advantages of client-side processing. This extension permits direct interaction with the BioImage Model Zoo through the chatbot, leveraging local computational resources for image analysis without compromising data privacy or security.

The chatbot's extension framework is designed to be highly adaptable, and capable of accommodating any valid Python function for diverse image analysis tasks. This versatility allows the chatbot to serve as a robust platform for custom automated analysis workflows, extending its utility to microscope control, pipeline management, AI model training, and system monitoring.

## Extension for Retrieval Augmented Generation

The BioImage.IO Chatbot has been augmented with a document retrieval extension, employing Retrieval Augmented Generation (RAG) to facilitate easy access to community-contributed software documentation and databases. This functionality is structured around a manifest YAML file that compiles metadata from a broad spectrum of documentation sources. Contributors are enabled to detail key information, including the name of the tool, a descriptive summary, and hyperlinks to the documentation, which can vary from GitHub repository markdown files to direct PDF document links.

Following the integration of this metadata, the chatbot proceeds to download and segment the documentation into smaller, digestible pieces. These segments are then processed using a text embedding model from OpenAI, generating text embeddings that capture the semantic essence of each documentation piece. These embeddings, in conjunction with the original text segments, are stored in a FAISS (Facebook AI Similarity Search) vector database, setting the stage for semantic retrieval operations within what is referred to as the "docs extension".

The core of this extension lies in its retrieval function, which serves as a chatbot extension tool. It converts text queries into embeddings with the aid of OpenAI text embedding models (), and FAISS then uses these embeddings to pinpoint and return the most relevant text segments based on cosine similarity measures. This mechanism ensures that users can access highly relevant information snippets that directly correspond to their queries.

The docs extension encompasses a variety of tools, each linked to distinct information sources, including specific software documentation like deepImageJ, and online resources such as the ImageJ wiki or bio.tools. Additionally, to bolster the chatbot's role in educational applications, a curated selection of bioimage analysis literature has been incorporated in a similar vein to the documentation extension. This enhancement significantly enriches the chatbot's capacity to support educational inquiries and learning-related interactions.

## Extension for Web Search

The BioImage.IO Chatbot has been equipped with a web search extension to broaden its search capabilities beyond specialized document access and database queries, facilitating internet searches. This extension integrates with the DuckDuckGo search engine to generate and execute search queries based on user inputs, aiming to retrieve succinct summaries from relevant websites.

Initiating with a call to DuckDuckGo's API, the chatbot utilizes keywords derived from the user's query to identify and list relevant web pages. The URLs obtained in this process lead to the fetching of HTML content from these pages, which is subsequently cleansed of markup tags to produce plain text. This prepared text is then segmented into smaller chunks, mirroring the approach taken in our Retrieval Augmented Generation (RAG) framework.

Following segmentation, each text chunk is processed using a text embedding model to generate corresponding embeddings. These embeddings, along with their text counterparts, are temporarily housed in a vector database, such as Chroma (https://www.trychroma.com/), for further processing.

The process continues with the execution of queries against this temporary vector database, aiming to condense the webpage text into essential content. This step is designed to refine the search results, ensuring that the summaries returned to the user are both relevant and concise. Ultimately, this extension furnishes users with a selection of brief, pertinent summaries, significantly enhancing the chatbot's search functionality and user experience by providing detailed and relevant responses to broad internet queries.

## Extension for Running AI Models via BioEngine

To enable image analysis on the user's own data, we developed a code interpreter extension powered by Pyodide, which is a Python runtime for the web browser. The extension works by executing Python code generated by the chatbot, and allowing users to seamlessly load, preprocess images, and run AI models directly within the chat interface. The code interpreter has access to locally mounted folder and can use standard scientific libraries such as numpy, scipy, scikit-image in Python to perform image processing.

To enable advanced image analysis powered by AI models, the BioImage Model Zoo (https://bioimage.io), the BioEngine platform (https://github.com/bioimage-io/bioengine) has been integrated within the code interpreter to incorporate advanced AI capabilities into bioimage analysis, facilitating the access to state-of-the-art AI models for cloud-based execution. BioEngine is engineered to simplify the model execution process, removing the burden of managing dependencies and hardware requirements from the user. It allows for the submission of images in the form of numpy arrays via Python, utilizing the Nvidia Triton Inference Server for executing selected AI models and delivering analysis results.

## Extension for Controlling Microscopes

To advance the integration of hardware control within the realm of smart microscopy, we demonstrate hardware interaction with the microscope control extension. For demonstration, this extension guides the chatbot to generate control commands for the Squid[20] microscope using the squid control library.

For the implementation, we have created several key functions, including stage movement and image acquisition, which are implemented as hypha services in line with our chatbot's extension mechanism. Each of these tool functions is annotated with a JSON schema that provides a clear description and the function signature, thereby seamlessly integrating with the chatbot. By demonstrating hardware control functionalities, we aim to expand the use of the chatbot in supporting data acquisition in bioimaging, aiming for next generation smart microscopy applications, offering researchers an innovative, automated approach to conduct their bioimaging experiments efficiently.

## Extension for Vision-Based Inspection

To enhance the chatbot's capability for direct interaction with user-provided images and the output from executed image analysis code, we developed the Vision Inspector extension, leveraging OpenAI GPT-4's vision capabilities. This extension establishes a feedback loop, enabling the use of LLM-generated code for image analysis, execution within the code interpreter, and subsequent visual inspection of the outputs. It serves a multifaceted role, aiding in the examination of input data, offering analysis recommendations, amending code based on visual feedback during analysis, and interpreting results and plots. Internally, the extension utilizes the GPT-4 vision model ( model ID: gpt-4-vision-preview) to interpret images alongside queries, enriched with additional contextual information. For consistent visualization, images are displayed using matplotlib, ensuring uniform sizing and inclusion of axes for conveying image dimensions. Additionally, titles and multiple panels facilitate comparative inspections, enhancing the analytical and educational utility of the chatbot by providing insightful visual feedback.

# Supplementary information

**Supplementary Video 1: Demonstrating User Interactions with Bioimage Documentation and Databases**

This video highlights the BioImage.IO Chatbot's interaction with extensive bioimaging resources, showcasing its ability to respond to user inquiries with detailed information. This video features a user querying the chatbot about specific details retrieved from the BioImage Model Zoo documentation, exploring protein and cell images through the Human Protein Atlas, and locating studies with corresponding cell images in the BioImage Archive. Through these interactions, the video illustrates the chatbot's proficiency in accessing and synthesizing information from specialized databases to provide users with accurate and relevant data, enhancing the research and discovery process in bioimaging.

**Supplementary Video 2: Demonstration of AI-Powered Bioimage Analysis Using the BioImage.IO Chatbot**

This video showcases the use of the BioImage.IO Chatbot to perform AI-based bioimage analysis entirely within a user's browser, highlighting the local processing of data without server-side involvement. The sequence begins with the user loading a folder from their local file system. Upon the query "display the image," the chatbot generates Python code using image.io and matplotlib to parse and display the image. Following this initial display, the user commands "segment the image with Cellpose." The chatbot responds by consulting the BioEngine documentation to list available models and retrieve usage information for the Cellpose model. It then generates Python code to segment cells using Cellpose via the BioEngine. The first attempt results in an image array shape error, which the chatbot detects and corrects automatically. The video concludes by illustrating how users can command the chatbot to count objects and plot their size distribution, demonstrating the chatbot's error handling and interactive capabilities in executing complex image analysis tasks.